# A New Validity Index for Fuzzy-Possibilistic C-Means Clustering

*Mohammad Hossein Fazel Zarandi*[a,*], *Shahabeddin Sotudian*[a], *Oscar Castillo*[b]

[a]*Department of Industrial Engineering and Management Systems, Amirkabir University of Technology, Tehran, Iran*

[b]*Tijuana Institutes of Technology, Tijuana, Mexico*

**Abstract**

In some complicated datasets, due to the presence of noisy data points and outliers, cluster validity indices can give conflicting results in determining the optimal number of clusters. This paper presents a new validity index for fuzzy-possibilistic c-means clustering called Fuzzy-Possibilistic(FP) index, which works well in the presence of clusters that vary in shape and density. Moreover, FPCM like most of the clustering algorithms is susceptible to some initial parameters. In this regard, in addition to the number of clusters, FPCM requires a priori selection of the degree of fuzziness ($m$) and the degree of typicality ($\eta$). Therefore, we presented an efficient procedure for determining an optimal value for $m$ and $\eta$. The proposed approach has been evaluated using several synthetic and real-world datasets. Final computational results demonstrate the capabilities and reliability of the proposed approach compared with several well-known fuzzy validity indices in the literature. Furthermore, to clarify the ability of the proposed method in real applications, the proposed method is implemented in microarray gene expression data clustering and medical image segmentation.

**Highlights:**

- A novel validity index for fuzzy-possibilistic c-means clustering is proposed.
- The proposed index is more robust to noise than its counterparts in the literature.
- It considers the shape and density of clusters using the properties of the fuzzy-possibilistic covariance matrix.
- A new algorithm for determining the best initial parameters of FPCM clustering is presented.
- The experimental results using artificial and well-known datasets show that the proposed index outperforms its fuzzy counterparts.

## 1. Introduction

Clustering is an unsupervised pattern classification method that determines the intrinsic grouping in a set of unlabeled data. There are a great number of algorithms for clustering, based on crisp [1], probabilistic [2], fuzzy [3], and possibilistic methods [4]. The hard-clustering methods restrict each point of the data set to exactly one cluster. However, since Zadeh introduced the notion of fuzzy sets that produced the idea of allowing to have membership functions to all clusters [5], fuzzy clustering has been extensively applied in varied fields of science, such as engineering and medical sciences [6, 7, 8].

In clustering algorithms, there are no predefined classes; as a result, we need to determine the optimal or near-optimal number of clusters before clustering. In this regard, compactness and separation are two measures for the clustering assessment and selection of an optimal clustering scheme [9]. The closeness of cluster elements represents compactness and isolation between clusters indicates separation.

So far, a considerable number of validity indices have been developed to evaluate the clustering quality (see section 2)**.** In these approaches, to find the optimal or near-optimal number of clusters, clustering algorithms should be executed several times for each cluster number and its outputs should be implemented into the cluster validity index in order to find the optimal or near-optimal number of clusters. Thus, to achieve an optimal prototype using a validity index two conditions are unavoidable:

1- An algorithm that can find the best initial parameters of the clustering algorithm.

2- A validity function for assessing the worthiness of cluster schemes for the various number of clusters.

Once these two necessities are met, the strategy of finding an optimal number of clusters is straightforward: find the best initial parameters; then use the validity function to choose the best number of clusters.

All the clustering algorithms are susceptible to some initial parameters. For example, FCM may give various clustering results with a varied degree of fuzziness. Therefore, even though the number of clusters is given, these algorithms may give different results for the optimal number of clusters. In the current study, we use FPCM clustering instead of FCM and its fuzzy counterparts, and we will discuss the reason for this shortly. Therefore, to satisfy the first condition, we propose a new algorithm for determining the best initial parameters of FPCM clustering including the degree of fuzziness ($m$) and typicality ($\eta$). Firstly, the algorithm reconstructs the original dataset from the outputs of the FPCM algorithm for different amounts of $m$, $\eta$, and the number of clusters. Then, the differences between the predicted dataset and the original dataset are determined using the root mean squared error (RMSE). Finally, the best amounts of $\eta$ and $m$ are obtained by minimizing the cumulative root mean square error (CRMSE) for every pair of $(m, \eta)$.

For the second condition, we propose a novel validity index for fuzzy-possibilistic c-means clustering called FP index. The major difficulty for measuring the compactness of a validity index is significant variability in the density, shape, and the number of patterns in each cluster. To tackle this problem, we assess the dispersion of the data for each cluster and consider the shape and density of clusters using the properties of the fuzzy-possibilistic covariance matrix as a measure of compactness. Also, an essential characteristic of a validity index is its capability to handle noise and outliers. Since FCM and cluster validity indices designed on its basis are quite susceptible to noise, we use FPCM instead of FCM and its fuzzy counterparts. Moreover, we use a fuzzy-possibilistic exponential-type separation in the separation part of the proposed FP index because an exponential operation is extremely effective in dealing with Shannon entropy [10].

The proposed framework is one of the very first fuzzy-possibilistic approaches in the literature. In the forthcoming sections, using artificial and well-known datasets, capabilities of the proposed approach will be tested and then it would be implemented for clustering several real microarray datasets and medical images.

The remainder of this paper is organized as follows. The next section reviews several cluster validity indices and we also will discuss their advantages and disadvantages. A new cluster validity index is then proposed for fuzzy-possibilistic clustering in section 3. A method for the determination of the parameters of the proposed index is presented in section 4. Section 5 gives the comparisons of experimental results on a variety of datasets and the proposed method will be implemented in microarray gene expression data clustering and medical image segmentation. Finally, conclusions are presented in section 6.

## 2. Background

### 2.1. Fuzzy-Possibilistic C-Means clustering

Fuzzy C-Means (FCM) clustering and its variation are the most renowned methods in the literature. FCM was proposed at first by Dunn [11], and then generalized by Bezdek [3]. A disadvantage of the FCM clustering algorithm is that it is susceptible to noise. To attenuate such an effect, Krishnapuram and Keller eliminated the membership constraint in FCM and proposed the Possibilistic C-Means (PCM) algorithm [4]. The superiority of PCM is that it is extremely robust in the presence of outliers. However, PCM has several defects, i.e., it considerably relies on a good initialization and has the undesirable propensity to generate coincident clusters [12].

To address these shortcomings, Pal and Bezdek defined Fuzzy-Possibilistic C-Means clustering (FPCM) that merges the attributes of both FCM and PCM. FPCM overcomes the noise susceptibility of FCM and also resolves the coincident clusters problem of PCM. They believed that typicalities and memberships are indispensable for defining the accurate feature of data substructure in the clustering problem. In this regard, they defined the objective function of FPCM as follows [13]:

$$\min_{(U,T,V,X)} \left\{ J_{FPCM}(U,T,V,X) = \sum_{i=1}^{c} \sum_{j=1}^{N} (t_{ij}^{\eta} + u_{ij}^{m}) D^2(x_j, v_i) \right\}, \qquad (1)$$

with the following constraints:

$$\begin{cases} \sum_{i=1}^{c} u_{ij} = 1 & \forall j \in (1,2,\ldots,N) \\ \sum_{j=1}^{N} t_{ij}^{\eta} = 1 & \forall i \in (1,2,\ldots,c) \end{cases}, \qquad (2)$$

where $X = \{x_1, x_2, \ldots, x_N\} \subseteq \mathbb{R}^d$ is the dataset in $d$-dimensional vector space, $u_{ij}$ is the degree of belonging of the $j^{th}$ data to the $i^{th}$ cluster, $V = \{v_1, v_2, \ldots, v_c\}$ is the prototypes of clusters, $D(x_j, v_i)$ is the distance between the $j^{th}$ data and the $i^{th}$ cluster center, $m$ is the degree of fuzziness, $t_{ij}$ is the typicality, $U$ and $T$ are fuzzy and possibilistic partition matrices, respectively. $\eta$ is a suitable positive number, $c$ is the number of clusters, and $N$ is the number of data. This objective function can be solved via an iterative procedure where the degrees of membership, typicality, and the cluster centers are updated via [13]:

$$u_{ij} = \left(\sum_{k=1}^{c} \left(\frac{D(x_j, v_i)}{D(x_j, v_k)}\right)^{2/(m-1)}\right)^{-1}, 1 \leq i \leq c, \quad 1 \leq j \leq N, \tag{3}$$

$$t_{ij} = \left(\sum_{k=1}^{N} \left(\frac{D(x_j, v_i)}{D(x_j, v_k)}\right)^{2/(\eta-1)}\right)^{-1}, 1 \leq i \leq c, \quad 1 \leq j \leq N, \tag{4}$$

$$v_i = \frac{\sum_{k=1}^{N}(t_{ik}^{\eta} + u_{ik}^{m})x_k}{\sum_{k=1}^{N}(t_{ik}^{\eta} + u_{ik}^{m})}, 1 \leq i \leq c. \tag{5}$$

### 2.2. Validity Indices for fuzzy clustering

In this subsection, we review some methods for quantitative assessment of the clustering results, known as cluster validity methods. According to the work of Wang and Zhang [14], these methods can be grouped into three main categories:

1. Indices comprising only the membership values,

2. Indices comprising the membership values and dataset,

3. Other approaches

The earliest validity indices for fuzzy clustering, the partition coefficient $V_{PC}$, and the partition entropy $V_{PE}$, were introduced by Bezdek [15]. These indices are examples of the indices comprising only the membership values. Their essential drawback is the lack of connection to the geometrical structure [14]. Some researchers considered fuzzy memberships and the data structure to resolve this disadvantage. In the current manuscript, we will compare the performance of the proposed validity index with fifteen popular cluster validation indices in the literature. Table 1 lists these cluster validity indices. In this table, $x_j$ is the $j^{th}$ data point, $c$ is the number of clusters, $v_i$ are cluster centers, $u_{ij}$ is the degree of belonging of the $j^{th}$ data to the $i^{th}$ cluster and $N$ is the total number of patterns in a given data set. The last three indices in this table are based on general type 2 fuzzy logic. Higher-order fuzzy clustering algorithms are very well suited to deal with the high levels of uncertainties present in the majority of real-world applications. However, the immense computational complexity associated with such clustering algorithms has been a great obstacle for the practical applications [16].

Now, we focus our attention on a well-known index from the second category which is the partition coefficient and exponential separation index (PCAES) proposed by Wu and Yang [21]. $V_{PCAES}$ only utilizes membership values to validate the compactness measure and does not consider the structure of data, i.e., the relative distance between objects and cluster centers [9]. For this reason, it practiced weak in compactness measure. In order to tackle this problem, we use the fuzzy-possibilistic covariance matrix and membership values in the proposed compactness measure. In this way, we involve characteristics like density, shape, and patterns in the proposed index.

Additionally, $V_{PCAES}$ takes advantage of the exponential function to validate the separation measure, and also it involves the distance between the mean of cluster centers and cluster centers. The stimulus behind taking the exponential function is that an exponential operation is extremely effective in coping with Shannon entropy [27, 28] and Wu and Yang had asserted that an exponential-type distance gives a robust property. Nevertheless, the experimental results demonstrate that this index gives inappropriate results when the cluster centers are close to each other [9]. Figure 1 illustrates an example of the limited way in which $V_{PCAES}$ loses its capability to indicate the appropriate number of clusters. Intuitively, we know that there are 7 fuzzy clusters in this dataset. In Section 4, it will be demonstrated that $V_{PCAES}$ will detect four clusters. This problem occurs because $V_{PCAES}$ calculates the separation between clusters using only centroid distances. To tackle these problems in the proposed index, we use membership values and centroid distances to improve the separation measure.

What's more, a substantial feature of a validity index is its capability to handle noise and outliers. Because of the noise sensitivity of FCM, and the structure of compactness measure in $V_{PCAES}$; it is very susceptible to noises. To demonstrate the noise sensitivity of PCAES validity index, we considered a 5-clusters dataset and the optimum number of clusters obtained using $V_{PCAES}$ was 5.

**Table 1.** Fifteen well-known validity indices for fuzzy clustering.

| Name/Authors | Function | Ref. |
|---|---|---|
| Partition Coefficient | $\max\limits_{2\leq c\leq C_{max}} V_{PC}(U,V,X) = \frac{1}{n}\sum_{i=1}^{c}\sum_{j=1}^{N} u_{ij}^2$ | [15] |
| Partition Entropy | $\min\limits_{2\leq c\leq C_{max}} V_{PE}(U,V,X) = -\frac{1}{n}\sum_{i=1}^{c}\sum_{j=1}^{N} u_{ij}\log u_{ij}$ | [15] |
| Fukuyama and Sugeno | $\min\limits_{2\leq c\leq C_{max}} V_{FS}(U,V,X) = \sum_{i=1}^{c}\sum_{j=1}^{N} u_{ij}^m \|x_j-v_i\|^2 - \sum_{i=1}^{c}\sum_{j=1}^{N} u_{ij}^m \|v_i-\bar{v}\|^2$, $\quad \bar{v}=\frac{\sum v_i}{c}$ | [17] |
| Xie and Beni | $\min\limits_{2\leq c\leq C_{max}} V_{XB}(U,V,X) = \dfrac{\sum_{i=1}^{c}\sum_{j=1}^{N} u_{ij}^m\|x_j-v_i\|^2}{N.\min\limits_{i,j}\|v_i-v_j\|^2}$ | [18] |
| Kwon | $\min\limits_{2\leq c\leq C_{max}} V_{K}(U,V,X) = \dfrac{\sum_{i=1}^{c}\sum_{j=1}^{N} u_{ij}^2\|x_j-v_i\|^2 + \frac{1}{c}\sum_{i=1}^{c}\|v_i-\bar{v}\|^2}{\min\limits_{i\neq k}\|v_i-v_k\|^2}$, $\quad \bar{v}=\frac{\sum_{j=1}^{N} x_j}{N}$ | [19] |
| Gath and Geva | $\min\limits_{2\leq c\leq C_{max}} V_{FHV}(U,V,X) = \sum_{i=1}^{c}[\det(F_i)]^{\frac{1}{2}}\qquad F_i=\dfrac{\sum_{j=1}^{N}(u_{ij}^m)(x_j-v_i)(x_j-v_i)^T}{\sum_{j=1}^{N}(u_{ij}^m)}$ | [20] |
| Wu and Yang | $\max\limits_{2\leq c\leq C_{max}} V_{PCAES}(U,V,X) = \sum_{i=1}^{c}\sum_{j=1}^{N}\frac{u_{ij}^2}{u_M} - \sum_{i=1}^{c}\exp\left(-\min\limits_{i\neq k}\{\frac{\|v_i-v_k\|^2}{B_T}\}\right)$, $u_M=\min\limits_{1\leq i\leq c}(\sum_{j=1}^{N} u_{ij}^m)$, $B_T=\sum_{s=1}^{c}\frac{\|v_s-\bar{v}\|^2}{c}$, $\bar{v}=\sum_{j=1}^{N}\frac{x_j}{N}$ | [21] |
| Zhang et al. | $\min\limits_{2\leq c\leq C_{max}} V_W(U,V) = \dfrac{Var^N(U,V)}{Sep^N(c,U)}$ <br> $Var^N(U,V) = Var(U,V)/\max\limits_{c}(Var(U,V))\qquad Sep^N(U,V) = Sep(c,U)/\max\limits_{c}(Sep(c,U))$ <br> $Sep(c,U) = 1-\max\limits_{i\neq j}\left(\max\limits_{x_k\in X}\min(u_{ik},u_{jk})\right)\qquad Var(U,V) = \left(\sum_{i=1}^{c}\sum_{j=1}^{N} u_{ij}\, d^2(x_j,v_i)/n(i)\right)\times\left(\dfrac{c+1}{c-1}\right)^{1/2}$ | [22] |
| Rezaee | $\min\limits_{2\leq c\leq C_{max}} V_{SC}(c,U) = Sep(c,U)/\max\limits_{c}(Sep(c,U)) + Comp(c,U)/\max\limits_{c}(Comp(c,U))$ <br> $Comp(c,U)=\sum_{i=1}^{c}\sum_{j=1}^{N} u_{ij}^2\|x_j-v_i\|^2$, $Sep(c,U)=\dfrac{2}{c(c-1)}\sum_{p\neq q}\left[\sum_{j=1}^{N}(\min(u_{F_p}(x_j),u_{F_q}(x_j))\times h(x_j)\right]$, $h(x_j)=-\sum_{i=1}^{c} u_{F_p}(x_j)\log_a u_{F_q}(x_j)$ | [23] |
| Zhang et al. | $\max\limits_{2\leq c\leq C_{max}} V_{WGLI} = (2MMD+Q_B)/3$, <br> $MMD=\frac{1}{n}\sum_{j=1}^{N}\max\limits_{1\leq i\leq C} u_{ij}$, $Q_B=\sum_i(e_{ij}-a_ia_j)$, $j=\max\limits_{k}(e_{ik})$, $a_i=\sum_i e_{ij}=\frac{1}{2M}\sum_{i\in V_l}\sum_{j\in V} A(i,j)$ <br> $A$ is the adjacency matrix and $M$ is the number of edges in a bipartite network. | [24] |
| Fazel Zarandi et al. | $\max\limits_{2\leq c\leq C_{max}} V_{ECAS}(c) = \dfrac{EC_{comp}(c)}{\max\limits_{c}(EC_{comp}(c))} - \dfrac{ES_{sep}(c)}{\max\limits_{c}(ES_{sep}(c))}$ <br> $EC_{comp}(c)=\sum_{i=1}^{c}\sum_{j=1}^{N} u_{ij}^m\exp\left(-\left(\dfrac{\|x_j-v_i\|^2}{\beta_{comp}}+\dfrac{1}{c+1}\right)\right)\qquad \beta_{comp}=\dfrac{\sum_{k=1}^{N}\|x_k-\bar{v}\|^2}{n(i)}$ <br> $\bar{v}=\sum_{j=1}^{N}\frac{x_j}{N}\qquad n(i)$ is the number of data in cluster $i$ <br> $ES_{sep}(c)=\sum_{i=1}^{c}\exp\left(-\min\limits_{i\neq j}\left\{\dfrac{(c-1)\|v_i-v_j\|^2}{\beta_{sep}}\right\}\right)\qquad \beta_{comp}=\dfrac{\sum_{b=1}^{c}\|v_b-\bar{v}\|^2}{c}$ | [9] |
| Fazel Zarandi et al. | $\min\limits_{2\leq c\leq C_{max}} V_{FNT}(U,V,X) = \dfrac{2}{c(c-1)}\sum_{p\neq q}^{c} S_{rel}(A_p,A_q)$ <br> $S_{rel}(A_p,A_q)$ is the relative similarity between two fuzzy sets $A_p$ and $A_q$. | [25] |
| Askari et al. | $\min\limits_{2\leq c\leq C_{max}} V_{GPF1} = \sum_{i=1}^{c} R_i^r\left(\sqrt{\prod_{q=1}^{r}\lambda_{qi}}\right)^{-1} \Big/ \sum_{i=1}^{c}\sum_{j=1}^{N} u_{ij}^m$ <br> $\lambda_{qi}$ is $q^{th}$ eigenvalue of fuzzy covariance norm matrix | [26] |
| Askari et al. | $\max\limits_{2\leq c\leq C_{max}} V_{GPF2} = \dfrac{1}{c}\sum_{k=1}^{c}\sum_{i=1}^{c}\sum_{j=1}^{N}|u_{kj}-u_{ij}|^m$ | [26] |
| Askari et al. | $\min\limits_{2\leq c\leq C_{max}} V_{GPF3} = \left(c\sum_{i=1}^{c} R_i^r\left(\sqrt{\prod_{q=1}^{r}\lambda_{qi}}\right)^{-1}\right) \Big/ \left(\left(\sum_{i=1}^{c}\sum_{j=1}^{N} u_{ij}^m\right)\left(\sum_{k=1}^{c}\sum_{i=1}^{c}\sum_{j=1}^{N}|u_{kj}-u_{ij}|^m\right)\right)$ | [26] |

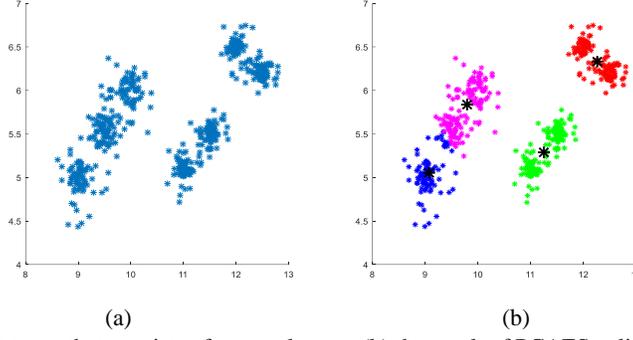

(a) (b)

**Figure 1.** (a) A data set that consists of seven clusters, (b) the result of PCAES validity index.

Additionally, we added 100 noisy points to the previous dataset and due to the noise sensitivity of $V_{PCAES}$; it can only detect four well-separated clusters in this noisy dataset. These datasets are depicted in Figure 2. To solve this problem in the proposed index, we use FPCM clustering instead of FCM or PCM clustering. FPCM clustering overcomes the noise susceptibility of FCM and also resolves the coincident clusters problem of PCM. In the next section, we will propose a new validity index for Fuzzy-Possibilistic C-Means clustering in order to overcome these shortcomings.

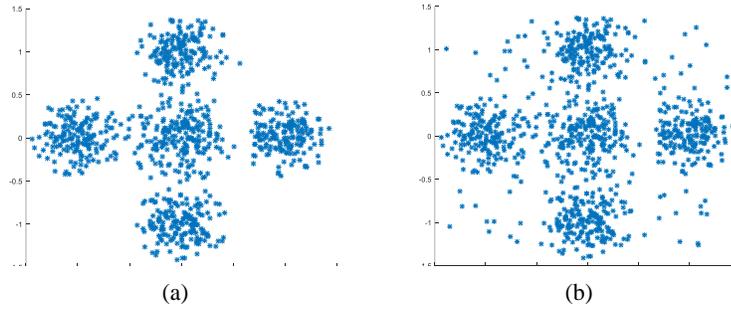

(a) (b)

**Figure 2.** (a) A dataset that consists of five clusters (b) Previous dataset + 100 noisy. points.

## 3. The proposed validity index

In the previous section, we reviewed the most widely used validity indices found in the literature, and we explained the disadvantages of some of these methods. Moreover, we suggested some solutions to address these issues. Now, we propose a new validity index for fuzzy-possibilistic c-means clustering which considers differences in cluster density, shape, and orientation and works well in the presence of noise. We will show that this validity index can effectively address these issues.

**Definition**: Let $X = \{x_1, x_2, \ldots, x_n | x_i \in \Re^p\}$ be a fuzzy-possibilistic c-partition of the dataset with $c$ cluster centers $v_i$, such that $V = \{v_1, v_2, \ldots, v_c\}$ and $u_{ij}$ as fuzzy membership of data point $x_j$ belonging to $i^{th}$ cluster and $t_{ij}$ as typicality of data point $x_j$ belonging to $i^{th}$ cluster.

The FP validity index has the following form:

$$V_{FP}(U,T,V,X) = Comp(c,U,T,V,X) + Sep(c,U,T,V), \qquad (6)$$

where, $Comp(c,U,T,V,X)$ is the compactness of the fuzzy-possibilistic c-partition which is defined as follows:

$$Comp(c,U,T,V,X) = \sum_{i=1}^{c} \frac{1}{trace(F_i)} \sum_{j=1}^{N} (t_{ij}^{\eta} + u_{ij}^{m}) \|x_j - v_i\|^2, \qquad (7)$$

where $m$ is degree of fuzziness, $\eta$ is degree of typicality and $F_i$ is the fuzzy-possibilistic covariance matrix of the $i^{th}$ cluster which is defined as follows:

$$F_i = \frac{\sum_{j=1}^{N}(t_{ij}^{\eta} + u_{ij}^{m})(x_j - v_i)(x_j - v_i)^T}{\sum_{j=1}^{N}(t_{ij}^{\eta} + u_{ij}^{m})}. \qquad (8)$$

In the compactness part, if dispersion within the cluster's augments, then clusters become less compact. Thus, the sum of fuzzy-possibilistic variations of clusters is an appropriate indication of the compactness of clusters. It worth mentioning that Equation 7 combines the advantages of fuzzy and possibilistic modeling with the power of the covariance matrix as a measure of compactness.

A significant obstacle for measuring the compactness is a considerable variation in the density, shape, and the number of patterns in each cluster. To tackle this problem, we can evaluate the variation of data for each cluster using the attributes of the fuzzy-possibilistic covariance matrix. In general, when compactness in a cluster is greater than the one of another cluster, the trace of that cluster covariance matrix will be less than the other. Owing to this inverse correlation between the trace of the cluster covariance matrix and compactness, we have put this term in the denominator to show this inverse correlation. Moreover, we use the trace of a matrix instead of its determinant because the computational complexity of computing the determinant is much greater than the complexity of trace.

$Sep(c, U, T, V)$ is fuzzy-possibilistic exponential-type separation of clusters which is defined as:

$$Sep(c, U, T, V) = \sum_{i=1}^{c} \sum_{j=1}^{N} (t_{ij}^{\eta} + u_{ij}^{m}) exp\left(-\min_{i \neq j}\left(\left(\frac{\|v_i - v_j\|}{\|v_i - \bar{v}\|}\right)^m\right)\right), \quad (9)$$

where $\bar{v} = \frac{\sum v_i}{c}$. The fuzzy-possibilistic exponential-type separation is similar to the exponential function of the separation measure in $V_{PCAES}$. According to research conducted by Wu and Yang, an exponential-type distance is more robust based on the influence function analysis [10]. Furthermore, we have combined the fuzziness and possibility in each row of $U$ and $T$ with the exponential-type separation. In section 2, we showed that $V_{PCAES}$ has inappropriate results when the cluster centers are close to each other. The experimental results indicate that FP index can correctly determine the number of clusters for this type of dataset (see section 4).

$Sep(c, U, T, V)$ and $Comp(c, U, T, V, X)$ have different scales, as a result: they need to be normalized before calculating $V_{FP}$. First, we explain each of them with respect to $c = 2,3,\ldots, c_{max}$ as follows:

$$Sep(c, U, T, V) = \{Sep(2, U, T, V), Sep(3, U, T, V), \ldots, Sep(c_{max}, U, T, V)\}, \quad (10)$$

$$Comp(c, U, T, V, X) = \{Comp(2, U, T, V, X), Comp(3, U, T, V, X), \ldots, Comp(c_{max}, U, T, V, X)\}. \quad (11)$$

For each measure, the maximum values are computed as:

$$Sep_{max} = \max_{c}(Sep(c, U, T, V)), \quad (12)$$

$$Comp_{max} = \max_{c}(Comp(c, U, T, V, X)). \quad (13)$$

Then, the normalized separation and compactness can be computed as:

$$Sep^N(c, U, T, V) = \frac{Sep(c, U, T, V)}{Sep_{max}}, \quad (14)$$

$$Comp^N(c, U, T, V, X) = \frac{Comp(c, U, T, V, X)}{Comp_{max}}. \quad (15)$$

Consequently, the proposed fuzzy-possibilistic cluster validity index $V_{FP}$ can be redefined as:

$$V_{FP}(U, T, V, X) = Comp^N(c, U, T, V, X) + Sep^N(c, U, T, V). \quad (16)$$

In the proposed validity index, a large value for the compactness measure over $c$ indicates a compact partition and, a large value for the separation measure over $c$ indicates well-separated clusters. Therefore, the optimum value of $c$ is obtained by maximizing $V_{FP}(U, T.V, X)$ over $c = 2,3,\ldots, c_{max}$.

What's more, the time and space complexity of validity indices depend on the underlying clustering algorithms. The time complexity of FCM and FPCM clustering algorithms is $O(tkNn^2)$ where $t, k, n$, and $N$ are the numbers of iterations, clusters, features, and objects, respectively. Additionally, the space complexity of these two algorithms is $O(Nn + kN + n^2)$. In addition, due to the fact that solving most of the common optimization formulations of clustering is NP-hard (in particular, solving the popular FCM and FPCM clustering problems), solving validity indices is also NP-hard.

## 4. A procedure for determining the parameters of the proposed method

In addition to the number of clusters, FPCM and its various extensions require a priori selection of the degree of fuzziness ($m$) and the degree of typicality ($\eta$). During the past few decades, various ranges and values for the optimum degree of fuzziness have been proposed. Here, we briefly review studies that have proposed a range or a method for determining the optimal degree of fuzziness. Then, we will present an efficient procedure for determining an optimal value for $m$ and $\eta$. Bezdek was one of the first scientists who introduced a heuristic procedure for finding an optimum value for $m$ [29]. McBratney and Moore [30] observed that the objective function value $J_m$ reduces monotonically with augmenting $c$ and $m$. Furthermore, they demonstrated that the greatest change in $J_m$, occurred around $m = 2$. Choe and Jordan [31] proposed an algorithm for finding the optimum $m$ using the concept of fuzzy decision theory. Yu et al. [32] defined two theoretical rules for selecting the weighting exponent in the FCM. According to their approach, they revealed the relationship between the stability of the fixed points of the FCM and the data set itself. Okeke and Karnieli [33] presented a procedure using the output of the fuzzy clustering. Their method predicts the original data using the idea of linear mixture modeling. The formula for reconstructing the original dataset has the following form:

$$\tilde{X} = \tilde{V}\tilde{U}, \tag{17}$$

where, $\tilde{X}$ is vector of predicted dataset. $\tilde{V}$ is vector of the FCM output centers and $\tilde{U}$ is the matrix of membership functions. Next, the differences between the predicted dataset and the original dataset are specified by the following formula [33]:

$$\sigma = \|X - \tilde{X}\|, \quad \sigma > 0, \quad \forall m \tag{18}$$

Eventually, the degree of fuzziness which corresponds to the minimum of $\sigma$ is the optimum value [33].

Since the amounts of $\eta$ and $m$ play an important role in FP index, we present an algorithm to tackle this problem. In the proposed algorithm, FPCM clustering is run for different values of $m$, $\eta$ and $c$. After that, the original dataset is reconstructed from the outputs of FPCM algorithm using the following formulas:

$$\tilde{X}^U = \tilde{V}\tilde{U}, \tag{19}$$
$$\tilde{X}^T = \tilde{V}\tilde{T}^N, \tag{20}$$

where, $\tilde{X}^U$ is the vector of the predicted dataset using membership functions matrix. $\tilde{U}$ is the matrix of membership functions for the FPCM algorithm and $\tilde{V}$ is the vector of the FPCM centers. $\tilde{X}^T$ is the vector of the predicted dataset using normalized typicality matrix. $\tilde{T}^N$ represents the normalized typicality matrix and is defined as:

$$\tilde{T}^N(c,N) = \frac{\tilde{T}(c,N)}{\sum_c \tilde{T}(c,N)}. \tag{21}$$

Then, the difference between the predicted dataset and the original dataset is determined by the root mean squared error (RMSE).

$$RMSE = \sqrt{\frac{\sum_{d=1}^{N}\sum_{i=1}^{c}(x_{id} - \tilde{x}_{id})^2}{N}}, \tag{22}$$

where, $x_{id}$ and $\tilde{x}_{id}$ are the actual and predicted datasets, respectively, $d$ is the number of data and $c$ is the number of clusters. Thus, $RMSE_{Total}$ can be defined as:

$$RMSE_{Total} = RMSE_T + RMSE_U, \tag{23}$$

where, $RMSE_T$ and $RMSE_U$ are the root mean squared errors computed by $\tilde{X}^T$ and $\tilde{X}^U$, respectively. After that, cumulative root mean square error (CRMSE) for every pair of $(m, \eta)$ is defined as:

$$CRMSE(m, \eta) = \sum_{c=2}^{c_{max}} RMSE_{Total}(m, \eta, c), \tag{24}$$

where, $c_{max}$ is the maximum number of clusters. Finally, an optimal value for $m$ and $\eta$ can be found by minimizing $CRMSE$ over $\eta$ and $m$. The steps of the proposed algorithm can be seen in Algorithm 1.

Algorithm 1 runs FPCM clustering and computes $V_{FP}$ with respect to $c = 2, 3, \ldots, c_{max}$. There is no universal agreement on what value to use for $c_{max}$. The value of $c_{max}$ can be selected in accordance with the user's knowledge about the dataset; however, as this is not always feasible, a lot of researchers use $c_{max} = \sqrt{N}$ [34]. What's more, the variation of FP index values for all experimental data sets demonstrates that the maximum of $V_{FP}$ exists between 2 and $\sqrt{N}$ (see Section 4). In order to show the behavior of Algorithm 1, the dataset which is shown in Fig 2 (b) is used as input data. Let $c_{max} = 8$, $m_{max} = 5$ and $\eta_{max} =$ 5 be the initial values for Algorithm 1(the theoretical rules proposed by Yu et al. are used in order to define $m_{max}$ and $\eta_{max}$). Table 2 shows the cumulative root mean square error of this dataset. The elements of this table are the degree of typicality ($\eta$) and the degree of fuzziness ($m$) as input variables and CRMSE as results.

For instance, for $m = 1.2$, $\eta = 2.2$, CRMSE is 6.316. According to Table 2, the suitable values of $m$ and $\eta$ can be found by $CRMSE\ (m^*, \eta^*) = \min_{\eta} \min_{m} (CRMSE)$. Therefore, the suitable values of $m$ and $\eta$ are 1.6 and 2.2, respectively. Finally, the optimal number of clusters obtained using Algorithm 1 is five with $m = 1.6$ and $\eta = 2.2$. Figure 3 shows the variation of the proposed index values with the number of clusters for this dataset.

---

**Algorithm 1:** The proposed algorithm for determining the suitable values of c, m and η

> Define the initial parameters:
> - Set $c=2$ and determine the maximum number of the clusters ($c_{max}$)
> - Set $m=1.1$ and determine the maximum value for the degree of fuzziness ($m_{max}$)
> - Set $\eta = 1.1$ and determine the maximum value for the degree of typicality ($\eta_{max}$)

For $c = 2$ to $c_{max}$
    For $m = 1.1$ to $m_{max}$
        For $\eta = 1.1$ to $\eta_{max}$

Compute fuzzy prototypes $\tilde{V}$, membership functions ($\tilde{U}$) and typicality ($\tilde{T}$) using FPCM algorithm.

Reconstruct the original dataset using Equation 19 and Equation 20.

Compute $RMSE_{Total}$, the difference between the original dataset and the predicted dataset using Equation 23.

End for

    End for

        End for

Compute $CRMSE$ for every pair of $(m, \eta)$ using Equation 24.

Find $m^*$ and $\eta^*$ such that $CRMSE\ (m^*, \eta^*) = \min_{\eta} \min_{m}(CRMSE)$.

Compute the FP index ($V_{FP}$) with $m = m^*$ and $\eta = \eta^*$ for $c = 2, 3, \ldots, c_{max}$

Determine the optimum value of $c$ by maximizing $V_{FP}$ over $c = 2, 3, \ldots, c_{max}$.

---

Table 2. CRMSE values for different $\eta$ and $m$.

| $\eta$ \ $m$ | 1.2 | 1.6 | 2 | 2.2 | 2.6 | 3 | 3.4 | 3.8 | 4.2 | 4.4 | 4.6 | 5 |
|---|---|---|---|---|---|---|---|---|---|---|---|---|
| 1.2 | 6.225 | 6.344 | 6.285 | 6.319 | 6.302 | 6.265 | 6.357 | 6.316 | 6.228 | 6.329 | 6.278 | 6.367 |
| 1.6 | 6.333 | 6.328 | 6.383 | 6.408 | 6.343 | 6.305 | 6.333 | 6.287 | 6.238 | 6.321 | 6.278 | 6.390 |
| 2 | 6.350 | 6.343 | 6.281 | 6.340 | 6.307 | 6.314 | 6.310 | 6.351 | 6.336 | 6.357 | 6.363 | 6.294 |
| 2.2 | 6.316 | 6.190 | 6.338 | 6.252 | 6.381 | 6.405 | 6.597 | 6.342 | 6.322 | 6.316 | 6.281 | 6.421 |
| 2.6 | 6.313 | 6.307 | 6.346 | 6.440 | 6.341 | 6.331 | 6.239 | 6.244 | 6.254 | 6.338 | 6.330 | 6.313 |
| 3 | 6.312 | 6.292 | 6.305 | 6.311 | 6.355 | 6.265 | 6.282 | 6.313 | 6.318 | 6.360 | 6.221 | 6.327 |
| 3.4 | 6.249 | 6.278 | 6.325 | 6.342 | 6.305 | 6.303 | 6.244 | 6.301 | 6.289 | 6.297 | 6.382 | 6.309 |
| 3.8 | 6.278 | 6.293 | 6.542 | 6.252 | 6.337 | 6.350 | 6.305 | 6.318 | 6.289 | 6.210 | 6.290 | 6.321 |
| 4.2 | 6.363 | 6.365 | 6.306 | 6.315 | 6.406 | 6.323 | 6.354 | 6.330 | 6.327 | 6.365 | 6.273 | 6.269 |
| 4.4 | 6.308 | 6.302 | 6.658 | 6.289 | 6.339 | 6.338 | 6.328 | 6.288 | 6.306 | 6.649 | 6.339 | 6.288 |
| 4.6 | 6.265 | 6.328 | 6.367 | 6.337 | 6.275 | 6.313 | 6.338 | 6.456 | 6.349 | 6.335 | 6.316 | 6.281 |
| 5 | 6.287 | 6.263 | 6.313 | 6.298 | 6.322 | 6.292 | 6.354 | 6.347 | 6.358 | 6.292 | 6.277 | 6.421 |

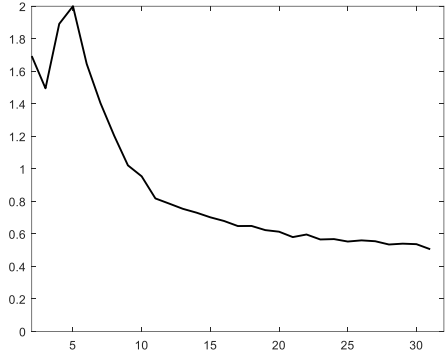
**Figure 3.** The variation of the proposed index values with the number of clusters

## 5. Experimental Results

In this section, to ascertain the effectiveness of FP index, we conducted comparisons between FP index and some well-known indices in the literature which we reviewed in section 2. In the next subsections, FP index will be evaluated using several synthetic and real-world datasets. Moreover, in order to clarify the ability of the proposed method in real applications, the proposed method is implemented in microarray gene expression data clustering and medical image segmentation. In the computational experiments, all the indices computed using the same input in order to achieve comparable results. In this regard, the clustering algorithm is run, and then the resulting $U$ matrix, the prototypes of clusters, and the other inputs needed for the indices are used for all the indices.

### 5.1. Artificial and real-world datasets

Eight artificial and five well-known datasets are considered for experiments. The eight artificial datasets are called Dataset_2_12, Dataset_2_5, Dataset_2_6, Dataset_2_7, Dataset_2_8, Dataset_2_10, Dataset_3_3, and Dataset_3_4. The names imply the number of clusters actually exists in the data and the dimension of data. For instance, in Dataset_2_5, there are five clusters and the dimension of the data is two. As can be seen, the artificial datasets include two and three-dimensional data where the number of clusters varies from three to twelve. These datasets are demonstrated in Figure 4. In addition, we will use six well-known datasets including Bupa Liver Disorder, Wine, Iris, Wisconsin Breast Cancer (WBC), Wisconsin Diagnostic Breast Cancer (WDBC), and Mammographic mass. These datasets are real-life data sets, which are freely accessible at [35]. The real-world datasets have dimensions in the range of four to thirty and the number of clusters varies from two to three.

We now present the experiment's results to compare $V_{FP}$ index with the other fifteen indices including $V_{PC}$, $V_{PE}$, $V_{FS}$, $V_{XB}$, $V_K$, $V_{FHV}$, $V_{PCAES}$, $V_W$, $V_{SC}$, $V_{WGLI}$, $V_{ECAS}$, $V_{FNT}$, $V_{GPF1}$, $V_{GPF2}$ and $V_{GPF3}$. In the proposed index, the optimum value of $c$ is obtained by maximizing $V_{FP}(U, T.V, X)$ over $c = 2,3, \ldots, c_{max}$. Figure 5 shows the variation of $V_{FP}$ with $c$ for all of the datasets. The maximum value of the index corresponds to the optimum number of clusters. These values for each data set can be found in Figure 5. For example, the proposed $V_{FP}$ index reaches the maximum ($V_{FP} = 2$) at $c^* = 2$ for the Iris data set, which properly reveals the underlying cluster number.
Furthermore, Table 3 summarizes the results obtained when the fifteen different validity indices were applied to the above-mentioned datasets. The column $c^*$ in Table 3 gives the actual number of clusters for each dataset, and other columns show the optimal cluster numbers obtained using each index. In this table, the highlighted entries correspond to the incorrect result of the indices.

As can be seen from this table, our validity index $V_{FP}$, $V_{GPF2}$ and $V_{GPF3}$, correctly recognize the correct number of clusters for all of the data sets. Here, it is worth mentioning that the general type 2 fuzzy clustering algorithms have outperformed the type 1 fuzzy clustering algorithms in many computational experiments [36]. This is due to the fact that a general type 2 fuzzy set offers a way to model higher levels of uncertainty because of additional degrees of freedom provided by its third dimension [36]. However, we should consider that the general type 2 fuzzy is computationally much more complex than type 1 fuzzy; particularly the defuzzifier process which is a very costly operation [36]. As a result, the immense computational complexity associated with general type 2 fuzzy clustering algorithms has been a great obstacle for practical applications. Therefore, although $V_{GPF2}$ and $V_{GPF3}$ perform with high accuracy, our proposed validity index achieved the same result with much less computational complexity.

In addition, $V_{FP}$, $V_{FHV}$, $V_W$, $V_{GPF1}$, $V_{GPF2}$ and $V_{GPF3}$ correctly recognize the number of clusters for datasets which the cluster centers are close to each other (Dataset_2_10 and Dataset_2_7). For noisy datasets (i.e. Dataset_2_6 and Dataset_2_5) the results showed that validity indices comprising only the membership values are very susceptible to noises, whereas some of the validity indices comprising the membership values and the dataset (i.e. $V_{FP}$ and $V_{FHV}$) are robust to noise.

### 5.2. Analysis of gene expression data

Microarray gene expression studies have been strongly followed by researchers over the last few years. The aim of these studies has been concentrated in finding the biologically considerable knowledge hidden under a large volume of gene expression data. In particular, recognizing gene groups that exhibit similar expression patterns (co-expressed genes) allows us to identify the set of genes involved in the same biological process; as a result, we can characterize unfamiliar biological facts. Clustering algorithms have shown an excellent capability for finding the underlying patterns in microarray gene expression profiles [37]. Given a set of genes, a clustering algorithm divides the genes into a number of distinct clusters based on certain similarity measures [38]. Each of the clusters corresponds to a specific macroscopic phenotype, such as clinical syndromes or cancer types [39]. In fact, a clustering algorithm should identify a set of clusters such that genes within a cluster possess high similarity as compared to the genes in different clusters and this task is not possible without knowing the optimal number of clusters.

In this subsection, three microarray gene expression datasets namely, Yeast sporulation, Rat CNS and Arabidopsis thaliana are tested and the capability of FP index will be analyzed from various perspectives. These datasets are adopted from [40, 41, 42]. For more details about the features of these datasets, please refer to [1].

After implementing the proposed validity index, the optimum number of clusters for Arabidopsis Thaliana and Yeast Sporulation datasets is 4 clusters and for Rat CNS dataset is 3 clusters. The variation of $V_{FP}$ with the number of clusters for these datasets are depicted in Figure 6. In fact, based on [40, 41, 42], the optimum number of clusters for Arabidopsis Thaliana and Yeast Sporulation datasets is 4 clusters and for Rat CNS dataset is 3 clusters. Therefore, the proposed validity index detected the correct number of clusters for all of these datasets. Table 4 summarizes the results obtained when the fifteen different validity indices were applied to these microarray gene expression datasets. As you can see, FP, GPF1, GPF2 and GPF3 indices correctly determine the number of clusters for all of these datasets. However, considering the high complexity of GPF1, GPF2, and GPF3 indices compared to the proposed index, we can conclude that FP index yields the best result for the gene expression datasets.

**Table 3.** The optimal number of clusters obtained by each cluster validity indices.

| Dataset | $c^*$ | PC | PE | FS | XB | K | FHV | PCAES | W | SC | WGLI | ECAS | FNT | GPF1 | GPF2 | GPF3 | FP |
|---|---|---|---|---|---|---|---|---|---|---|---|---|---|---|---|---|---|
| Dataset_2_12 | 12 | 10 | 13 | 7 | 12 | 12 | 12 | 12 | 12 | 12 | 12 | 12 | 12 | 12 | 12 | 12 | 12 |
| Dataset_2_5 | 5 | 4 | 2 | 5 | 4 | 4 | 5 | 4 | 4 | 4 | 4 | 4 | 4 | 4 | 5 | 5 | 5 |
| Dataset_2_6 | 6 | 7 | 7 | 6 | 6 | 6 | 7 | 6 | 7 | 6 | 6 | 6 | 7 | 7 | 6 | 6 | 6 |
| Dataset_2_7 | 7 | 2 | 2 | 7 | 6 | 7 | 7 | 4 | 7 | 6 | 7 | 6 | 6 | 7 | 7 | 7 | 7 |
| Dataset_2_8 | 8 | 2 | 2 | 7 | 7 | 8 | 8 | 8 | 8 | 8 | 8 | 8 | 8 | 8 | 8 | 8 | 8 |
| Dataset_2_10 | 10 | 10 | 10 | 5 | 8 | 5 | 10 | 5 | 10 | 10 | 8 | 10 | 10 | 10 | 10 | 10 | 10 |
| Dataset_3_3 | 3 | 2 | 2 | 3 | 2 | 2 | 3 | 2 | 3 | 3 | 3 | 2 | 2 | 3 | 3 | 3 | 3 |
| Dataset_3_4 | 4 | 2 | 2 | 4 | 4 | 4 | 4 | 4 | 4 | 4 | 4 | 4 | 4 | 4 | 4 | 4 | 4 |
| Liver Disorder | 2 | 2 | 2 | 4 | 2 | 2 | 18 | 2 | 2 | 2 | 2 | 2 | 2 | 2 | 2 | 2 | 2 |
| Wine | 3 | 2 | 2 | 13 | 3 | 3 | 3 | 3 | 3 | 3 | 3 | 3 | 2 | 3 | 3 | 3 | 3 |
| Iris | 2,3 | 2 | 2 | 5 | 2 | 2 | 3 | 2 | 2 | 2 | 2 | 2 | 2 | 2 | 2 | 2 | 2 |
| WBC | 2 | 2 | 2 | 12 | 2 | 2 | 2 | 2 | 2 | 2 | 2 | 2 | 2 | 2 | 2 | 2 | 2 |
| WDBC | 2 | 2 | 2 | 12 | 2 | 2 | 2 | 2 | 2 | 2 | 2 | 2 | 2 | 2 | 2 | 2 | 2 |
| Mammographic mass | 2 | 2 | 2 | 2 | 2 | 2 | 2 | 2 | 2 | 2 | 2 | 2 | 2 | 2 | 2 | 2 | 2 |

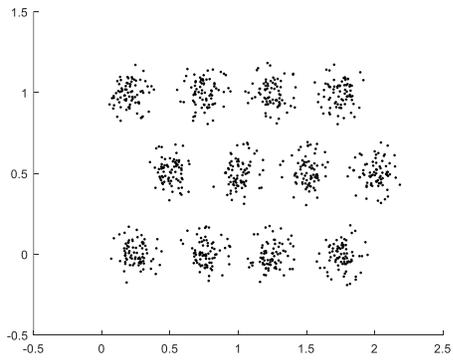

(a) Dataset_2_12

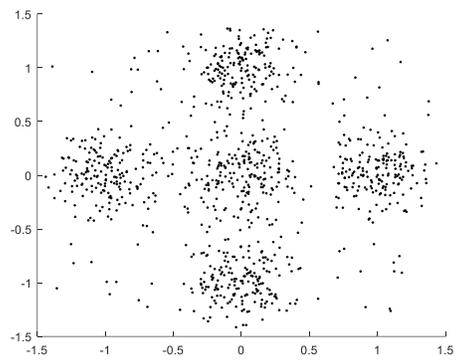

(b) Dataset_2_5

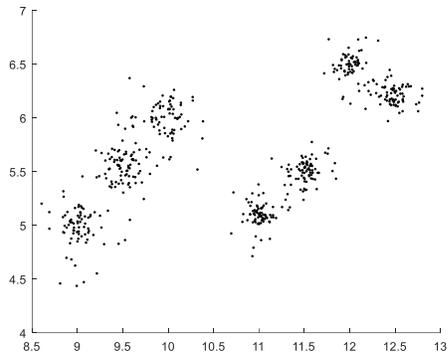

(c) Dataset_2_7

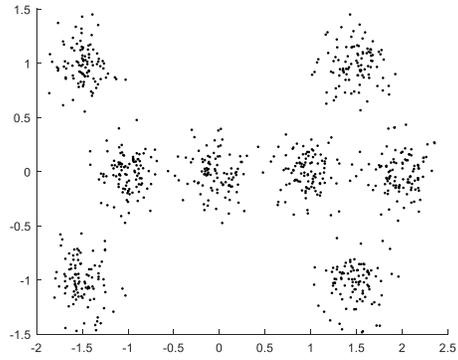

(d) Dataset_2_8

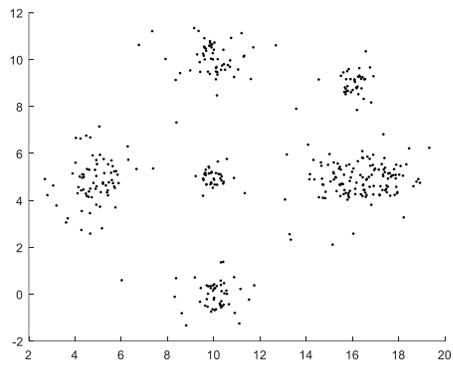

(e) Dataset_2_6

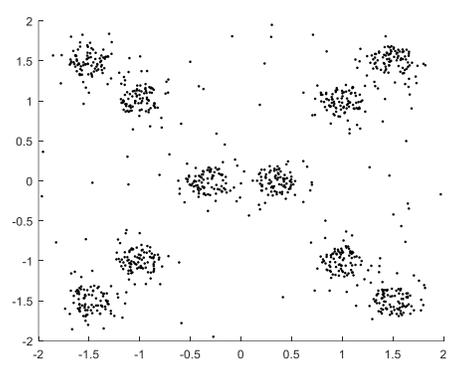

(f) Dataset_2_10

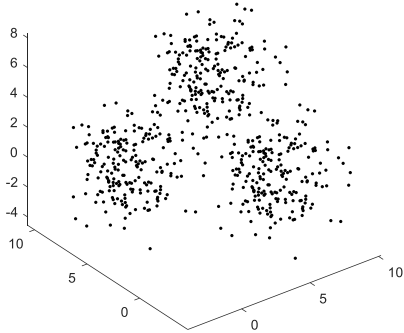

(g) Dataset_3_3

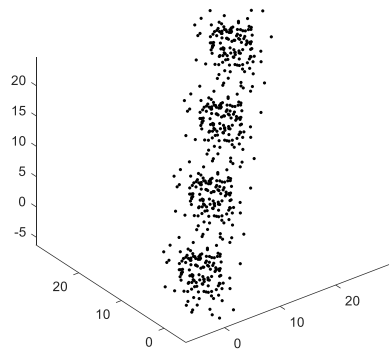

(h) Dataset_3_4

**Figure 4.** The eight artificial datasets

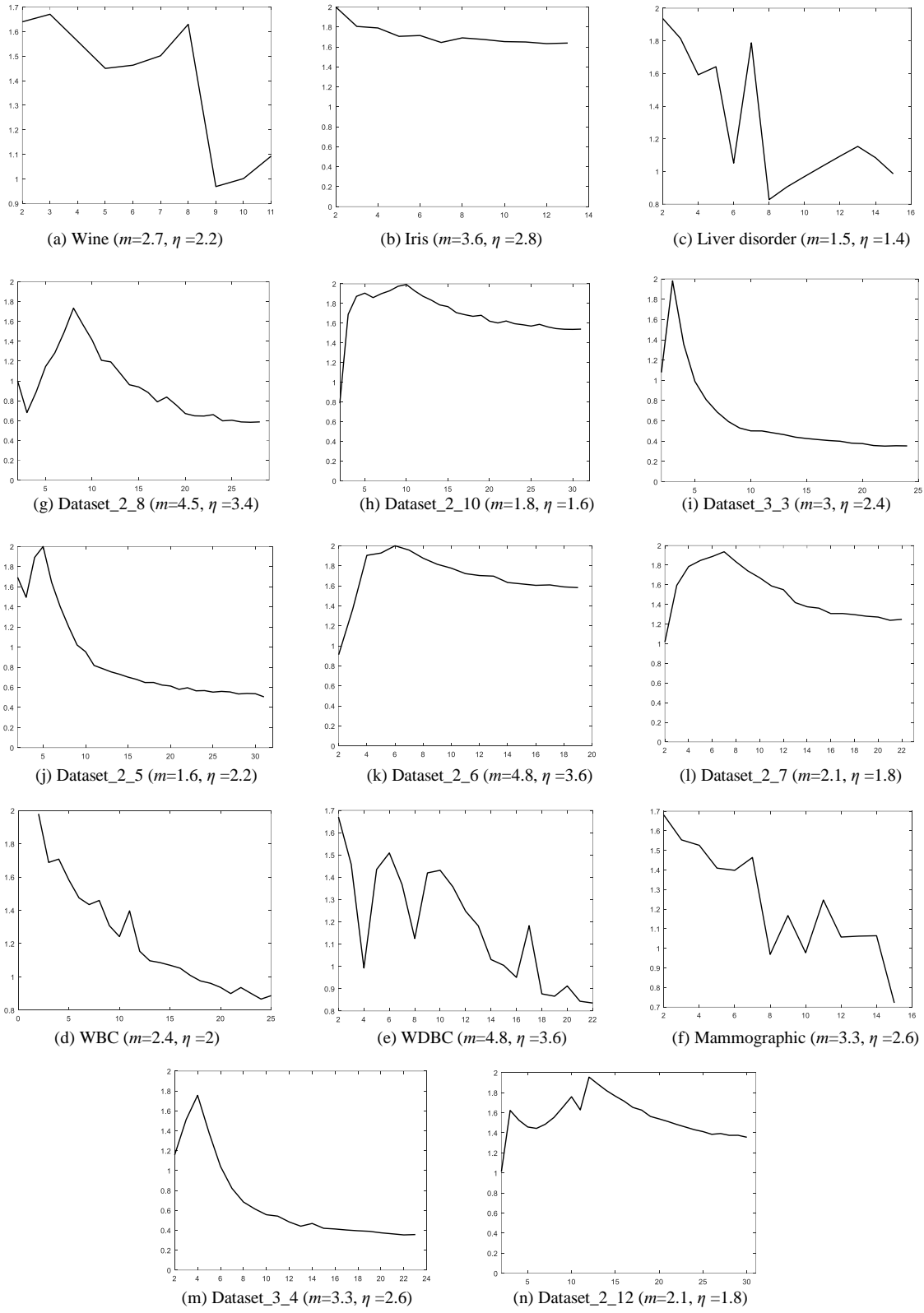

**Figure 5.** The variation of the proposed index values with the number of clusters for all data sets.

Moreover, towards a better understanding of the microarray gene expression context, we used Eisen plot as a visual tool. An Eisen plot is a two-dimensional false color plot that visualizes the expression levels of plenty of genes across several samples. Every row in the Eisen plot demonstrates a gene expression profile across the sample [43]. We have also generated a random sequence of genes for the simpler distinction between the effects of FP index and a random sequence of genes for each dataset. These plots are depicted in Figures 7–9. Lines in each Eisen plot are the boundaries of clusters. Here, due to the fact that the number of clusters has been determined properly, the genes within a cluster possess high similarity as compared to the genes in other clusters. Moreover, the genes in different clusters are properly separated.

It can be seen that FP index performs really well in determining the suitable number of clusters of the gene expression data sets. However, it should be noted that because of the complicated nature of the gene expression data sets, it is difficult to find a single partitioning that can be claimed to be the optimal partition. From the figures, it is apparent that the expression profiles of the genes of a cluster are similar to each other and they usually have similar color patterns. Moreover, these figures also demonstrate how the cluster profiles for the various groups of genes differ from each other, while the profiles within a group are similar.

**Table 4.** The optimal number of clusters obtained by each cluster validity indices

| Dataset | $c^*$ | PC | PE | FS | XB | K | FHV | PCAES | W | SC | WGLI | ECAS | FNT | GPF1 | GPF2 | GPF3 | FP |
|---|---|---|---|---|---|---|---|---|---|---|---|---|---|---|---|---|---|
| Rat CNS | 3 | **2** | **2** | 7 | **2** | **2** | 8 | **2** | **2** | 3 | 3 | **2** | **2** | 3 | 3 | 3 | 3 |
| Yeast | 4 | **2** | **2** | 3 | **2** | **2** | 7 | **2** | 4 | **3** | **2** | 4 | 4 | 4 | 4 | 4 | 4 |
| Arabidopsis | 4 | **2** | **2** | 7 | 4 | **3** | 7 | **2** | 4 | **2** | 4 | 4 | **3** | 4 | 4 | 4 | 4 |

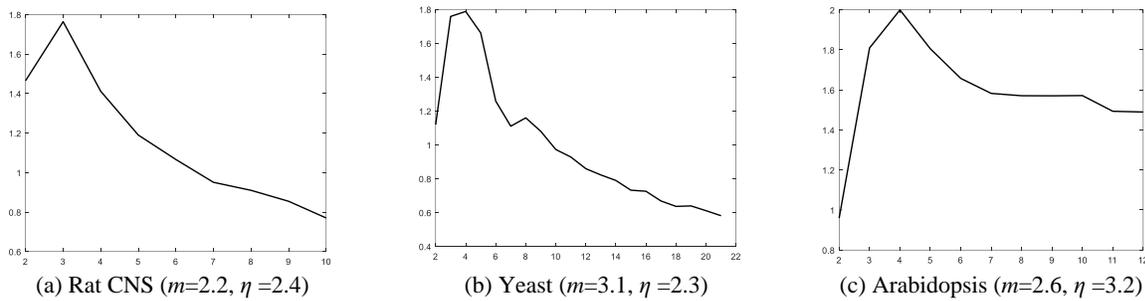

(a) Rat CNS ($m$=2.2, $\eta$=2.4)     (b) Yeast ($m$=3.1, $\eta$=2.3)     (c) Arabidopsis ($m$=2.6, $\eta$=3.2)

**Figure 6.** The variation of FP index values with the number of clusters for all microarray gene expression datasets.

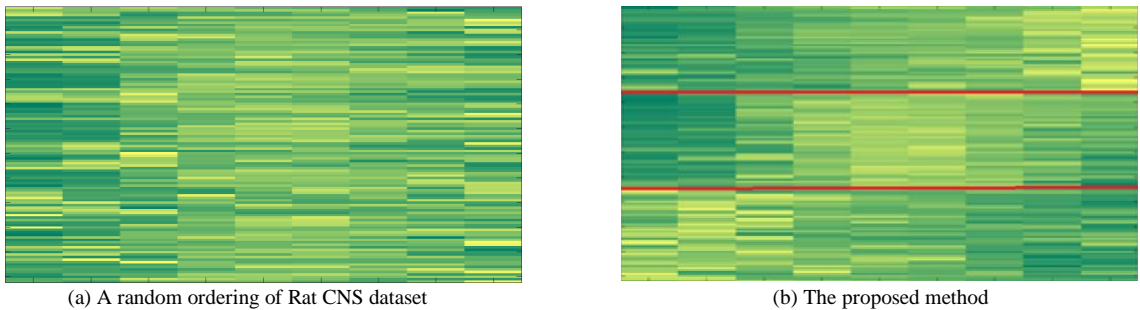

(a) A random ordering of Rat CNS dataset     (b) The proposed method

**Figures 7.** Eisen plot for Rat CNS dataset

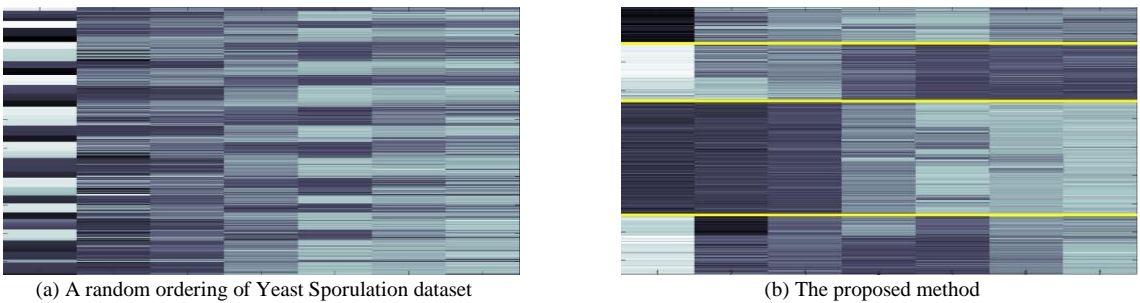

(a) A random ordering of Yeast Sporulation dataset     (b) The proposed method

**Figures 8.** Eisen plot for Yeast Sporulation dataset

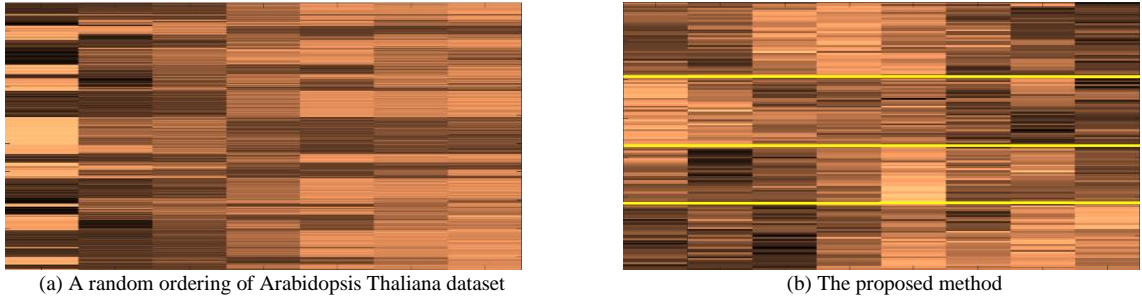

(a) A random ordering of Arabidopsis Thaliana dataset  (b) The proposed method

**Figures 9.** Eisen plot for Arabidopsis Thaliana dataset

### 5.3. Medical image segmentation

In this subsection, $V_{FP}$ is experimented on several liver images to show the applicability of this approach in medical image segmentation. In general, segmentation is the process of dividing an image into regions with similar properties such as color, brightness, texture, and contrast [1]. The existence of noise and low contrast in medical images are critical barriers that stand in the way of achieving a good segmentation system. Thus, in order to check the applicability of our method in this area, we use three medical images of the liver. For this purpose, two CT images and an ultrasound image of the liver are adopted from [44]. These medical data are shown in Figure 10.

The proposed index is performed for the datasets corresponding to these images, and the number of clusters is obtained four clusters for all of the datasets. The variation of FP index values with the number of clusters for these datasets are demonstrated in Figure 11. Moreover, the results of segmentation by FPCM clustering when the number of clusters is four are shown in Figure 12-14.

Table 5 summarizes the results obtained when the fifteen different validity indices were applied to the datasets corresponding to these images. As you can see, the segmentation results show that FP, ECAS, GPF1, GPF2 and GPF3 indices can successfully recognize the optimal number of clusters for all of these datasets. Consequently, FP index can effectively segment the cysts and lesions from the CT images and the ultrasound image, despite the gray level resemblance of adjoining organs and the various gray level of hepatic cysts and lesions in the images.

In general, the sensitivity of ultrasound images is significantly less than that of CT images in detecting lesions. Ultrasound images are not as detailed as those from CT or MRI scans. Its use is also limited in some parts of the body because the sound waves cannot go through the air (such as in the lungs) or through bone. In this experiment, we just want to show that our method is even able to detect the lesions in ultrasound images despite the gray level resemblance of adjoining organs and the various gray level of hepatic cysts and lesions in these images. As you can see in Figure 14, our method successfully detects two lesions. However, because of the low quality of this kind of images, our final result is not as good as CT images.

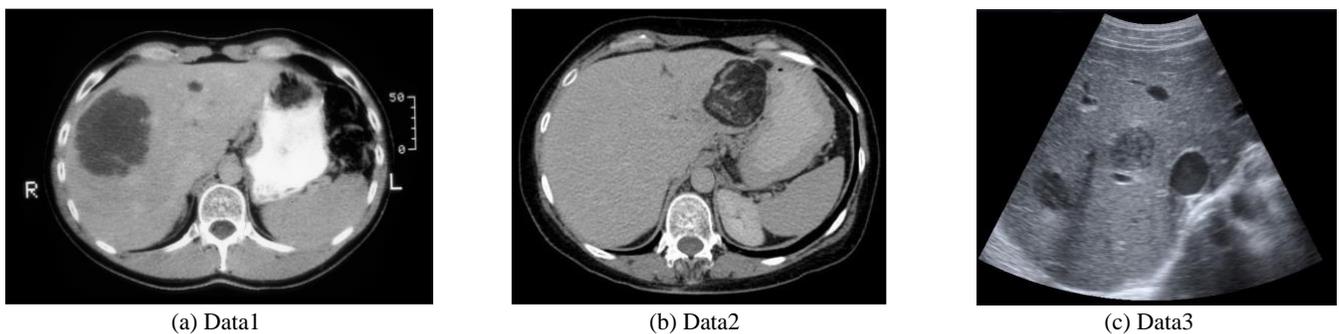

(a) Data1  (b) Data2  (c) Data3

**Figure 10.** (a) CT image reveals a large cystic lesion, (b) CT image demonstrates a large liver hepatic Angiomyolipoma, (c) Grey scale ultrasound shows two focal non-specific hypoechoic liver lesions.

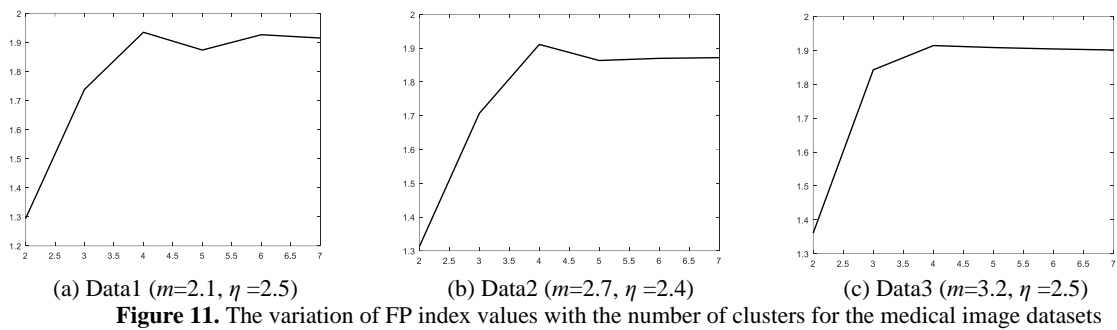

(a) Data1 (*m*=2.1, *η* =2.5)      (b) Data2 (*m*=2.7, *η* =2.4)      (c) Data3 (*m*=3.2, *η* =2.5)

**Figure 11.** The variation of FP index values with the number of clusters for the medical image datasets

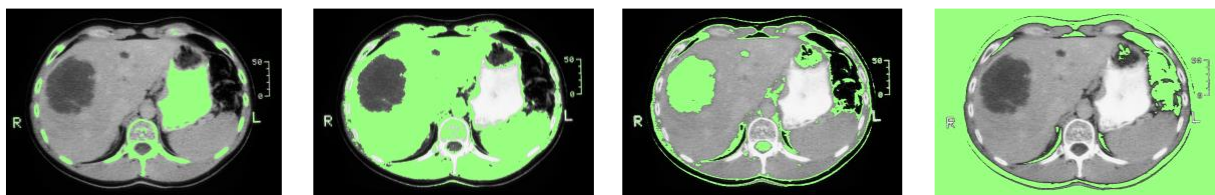

**Figure 12.** The segmented Data1

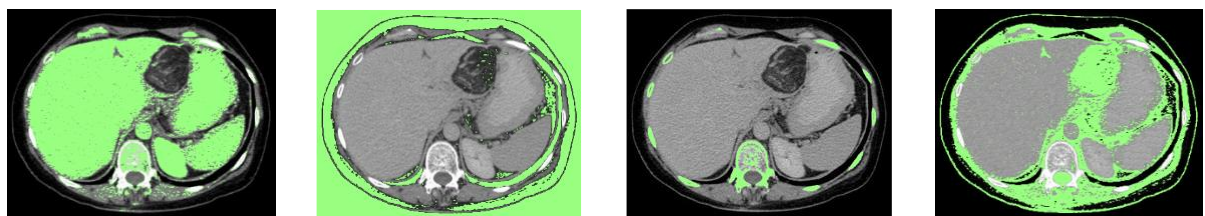

**Figure 13.** The segmented Data2

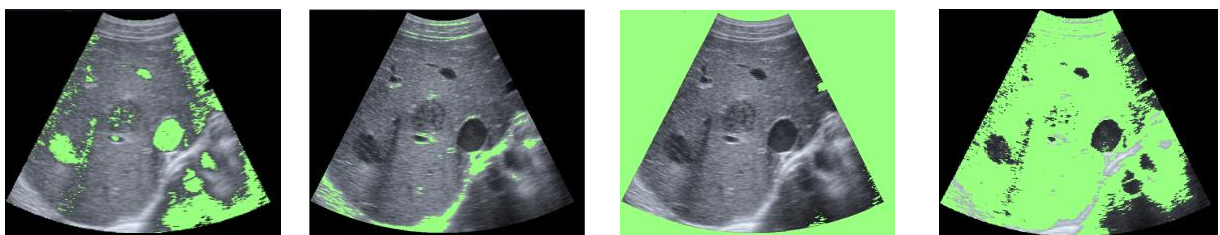

**Figure 14.** The segmented Data3

**Table 5.** The optimal number of clusters obtained by each cluster validity indices

| Dataset | PC | PE | FS | XB | K | FHV | PCAES | W | SC | WGLI | ECAS | FNT | GPF1 | GPF2 | GPF3 | FP |
|---|---|---|---|---|---|---|---|---|---|---|---|---|---|---|---|---|
| Data1 | 2 | 2 | 5 | 10 | 4 | 17 | 2 | 4 | 4 | 2 | 4 | 4 | 4 | 4 | 4 | 4 |
| Data2 | 2 | 2 | 5 | 5 | 2 | 4 | 4 | 8 | 4 | 4 | 4 | 4 | 4 | 4 | 4 | 4 |
| Data3 | 2 | 2 | 11 | 9 | 2 | 2 | 2 | 4 | 5 | 4 | 4 | 5 | 4 | 4 | 4 | 4 |

# 6. Conclusions

In this paper, we reviewed several fuzzy validity indices and discussed their advantages and disadvantages. We observed that some of them give incorrect results when the cluster centers are close to each other. Moreover, we demonstrated that most of these indices are susceptible to noise. To tackle these shortcomings, we proposed a new fuzzy-possibilistic validity index called FP index.

Furthermore, FPCM like other clustering algorithms is susceptible to some initial parameters. In this regard, in addition to the number of clusters, FPCM requires a priori selection of the degree of fuzziness ($m$) and the degree of typicality ($\eta$). Therefore, we presented an efficient procedure for determining the optimal values for $m$ and $\eta$.

For demonstrating the efficiency of FP index, we examined the proposed index using eight artificial and five well-known datasets. Our experiments demonstrated the effectiveness and flexibility of the FP validity index regarding sensitivity to cluster overlapping and difference in cluster shape and density in comparison with several well-known approaches in the literature. Moreover, we discussed the applications of the proposed approach in real microarray gene expression datasets and medical image segmentation. In both applications, we observed that our method has an excellent performance in determining the proper number of clusters, and is robust in the presence of noise.